\pdfoutput=1

\documentclass[11pt]{article}

\usepackage[]{ACL2023}

\usepackage{times}
\usepackage{latexsym}

\usepackage[T1]{fontenc}

\usepackage[utf8]{inputenc}

\usepackage{microtype}

\usepackage{inconsolata}

\usepackage{makecell}
\usepackage{graphicx}
\usepackage{siunitx}

\newcommand{\coo}{\ensuremath{\mathrm{CO_2}}}

\title{Efficient Large Language Models with Zero-Shot Adjustable Acceleration}

\author{Sajjad Kachuee \\
  Department of Electrical Engineering \\
  Sharif University of Technology \\
  Tehran, Iran \\
  \texttt{sajjad.kachuee@gmail.com} \\\And
  Mohammad Sharifkhani \\
  Department of Electrical Engineering \\
  Sharif University of Technology \\
  Tehran, Iran \\
  \texttt{msharifk@sharif.edu} \\}

\begin{document}
\maketitle
\begin{abstract}
Using Large Language Models (LLMs) in real-world applications presents significant challenges, particularly in balancing computational efficiency with model performance. Optimizing acceleration after fine-tuning and during inference is critical for building efficient architectures. This paper introduces Zero-Shot Adjustable Acceleration, a novel training and inference method that dynamically adjusts hardware utilization during inference without requiring additional fine-tuning. The proposed approach is applied to recent LLMs and evaluated across multiple classification and text generation tasks. Experimental results demonstrate that the method supports a wide range of zero-shot acceleration and achieves up to 11× speedup compared to the baseline.
\end{abstract}


\section{Introduction} \label{introduction}
In recent years, numerous large language models (LLMs) have emerged, with performance generally improving as model size increases. However, the development and deployment of these large-scale models are often limited to major corporations and research institutions due to their substantial computational and hardware requirements. This restricts their practical use in real-world applications such as industrial safety systems and time-sensitive data monitoring platforms, where computational efficiency is critical.

Training state-of-the-art LLMs demands vast amounts of data, energy, and hardware. For instance, LLaMA 3 by Meta AI \citep{grattafiori2024llama} required 39.3 million GPU hours on H100-80GB hardware, consuming 27.5 GWh of electricity and emitting an estimated 11,390 tons of \coo. Although Meta mitigates this impact through renewable energy and net-zero emissions, the resource intensity remains a significant barrier to broader adoption.

Inference latency and limited access to large datasets further complicate LLM deployment. While fine-tuning can adapt models to specific tasks, computational constraints continue to hinder their use in practical settings.

This work proposes a novel architecture and training policy that introduces a dynamic tuning parameter, enabling zero-shot adjustment of model acceleration. This approach promotes efficient resource utilization and reduces inference latency, particularly in services with fluctuating workloads, without requiring retraining. The main contributions of this work are:

\begin{itemize}
\item A new hidden activation pruning architecture that stably accelerates inference by up to 11×.
\item A fine-tuning policy that prepares the model for zero-shot adjustable acceleration, enabling flexible speed–accuracy trade-offs at inference time without additional training or re-finetuning.
\item A method that does not alter the core architecture and can be applied alongside other acceleration techniques.
\end{itemize}

In this paper, we refer to the hidden layer activations at a given position as the hidden activation. While influenced by the original embedding through residual connections, these activations are distinct from static word embeddings.

\section{Related Works}
This section reviews prior work on accelerating LLM inference, categorized into model compression, efficient architectures, and dynamic inference strategies.

\subsection{Model Compression Techniques}
Model compression aims to reduce computational complexity while minimizing accuracy loss. Approaches include weight compression \citep{yoo2023tf, nova2023gradient, belcak2023exponentially}, pruning attention heads, channels, layers, or blocks \citep{kwon2022fast, hu2023smartbert, xin2021berxit, sajjad2023effect}, and compressing hidden activations \citep{kim2022learned, lee2022sparse, anagnostidis2024dynamic, jaegle2021perceiver, kachuee2025latency}. Knowledge distillation trains smaller student models to mimic larger teachers \citep{sanh2019distilbert, jiao2019tinybert, sun2020mobilebert, xu2020deep}, while quantization reduces numerical precision to lower memory and speed up inference \citep{dettmers2022gpt3, dettmers2023qlora, guo2023lq}.

\subsection{Efficient Architecture Design}
Mixture-of-Experts (MoE) models scale efficiently by activating only a subset of experts per input \citep{sanseviero2023moe}. Sparse attention mechanisms, such as Longformer \citep{beltagy2020longformer} and BigBird \citep{zaheer2020big}, focus computation on relevant tokens. FlashAttention \citep{dao2022flashattention, dao2023flashattention, shah2024flashattention} optimizes attention computation, improving GPU utilization for long sequences. Other innovations include speculative decoding \citep{leviathan2023fast} and MEDUSA \citep{cai2401medusa}, enabling faster token prediction without accuracy loss.

\subsection{Dynamic Inference Techniques}
Dynamic inference methods adapt computation per token or layer. Early exit strategies, e.g., ELBERT \citep{xie2021elbert} and D-LLM \citep{jiang2024d}, terminate computation when confident predictions are achieved. RT-LM \citep{li2023rt} and Llumnix \citep{sun2024llumnix} manage uncertainty and load dynamically. Token-level routing \citep{huang2024harder, dotzel2024radial, hu2024blockllm} and FLOP allocation \citep{raposo2024mixture, malla2024copal} optimize computation across layers. Token Merging \citep{bolya2022token} and dynamic DNNs \citep{lou2021dynamic, qin2024dodo, kachuee2025latency, fu2024lazyllm} further accelerate inference by selectively pruning or merging activations and tokens.

Key challenges remain in maintaining stable inference performance, supporting diverse acceleration levels, and optimizing post-training efficiency.

\section{Proposed Method}
This section presents the proposed method. It begins with an overview of LLM architectures, highlighting key similarities and differences, followed by a description of attention layer functionality and the concept of hidden activation contribution within the attention mechanism. Next, the proposed hidden activation pruning strategy is introduced, detailing its methodology and benefits. The section concludes with an explanation of the zero-shot adjustable acceleration approach.

\subsection{LLMs Architectures Review} \label{sec3.1}
As noted in Section~\ref{introduction}, recent mainstream development of Large Language Models (LLMs) has primarily focused on decoder-only architectures. Following this trend, several research groups have introduced multiple LLM families, including LLaMA \citep{touvron2023llama}, Gemma \citep{team2024gemma}, and GPT \citep{radford2018improving}.

A review of various LLM architectures—such as LLaMA 2 \citep{touvron2023LLaMA2}, LLaMA 3 \citep{dubey2024llama}, Gemma 1 \citep{team2024gemma}, Gemma 2 \citep{team2024gemma2}, Mistral \citep{jiang2023mistral}, and GPT-2 \citep{radford2019language}—reveals substantial architectural similarities. These include overall structure, activation functions, residual branches, and attention layer designs. A comparative summary is provided in Appendix~\ref{llm_comparisons}.

\subsection{Hidden Activation Context Contribution}
The Transformer architecture \citep{vaswani2017attention} and its successors have achieved remarkable success across various Natural Language Processing (NLP) tasks. At the core of these models, the attention layer is the most critical component. This layer is primarily designed in two forms: bidirectional attention, commonly used in encoder architectures for text classification tasks, and causal attention, predominantly employed in decoder architectures for text generation tasks.

The key distinction between causal and bidirectional attention lies in how tokens interact during processing. Causal attention restricts each token to attend only to preceding tokens in the sequence, making it particularly suitable for autoregressive text generation. In contrast, bidirectional attention allows tokens to attend to both preceding and succeeding tokens, which is essential for tasks such as text classification and masked language modeling. The mathematical formulation of the attention layer is given by:

\begin{equation}
Att(Q,K,V) = \text{softmax}\Big(\frac{QK^T}{\sqrt{d_k}} + M\Big)V
\label{eq1}
\end{equation}

Here, $Q$ (queries), $K$ (keys), and $V$ (values) are matrices derived from the input context. The key dimension $d_k$ is used for scaling, preventing large dot-product values from dominating the softmax function. The attention mask $M$ differentiates between causal and bidirectional attention: in causal attention, $M$ is a lower triangular matrix with $-\infty$ in future positions to prevent tokens from attending to future inputs, whereas in bidirectional attention, $M = 0$, allowing all tokens to attend to each other.

The softmax function normalizes the attention scores, referred to as the soft-score, which is a square matrix with dimensions corresponding to the input sequence length. By multiplying the soft-score with the values matrix $V$, the contribution of each row in $V$—representing a single hidden activation—to the layer output is determined.

\subsection{Dynamic Resource Attention} \label{sec3.3}
\begin{figure}[t]
\centerline{\includegraphics[width=0.7\linewidth]{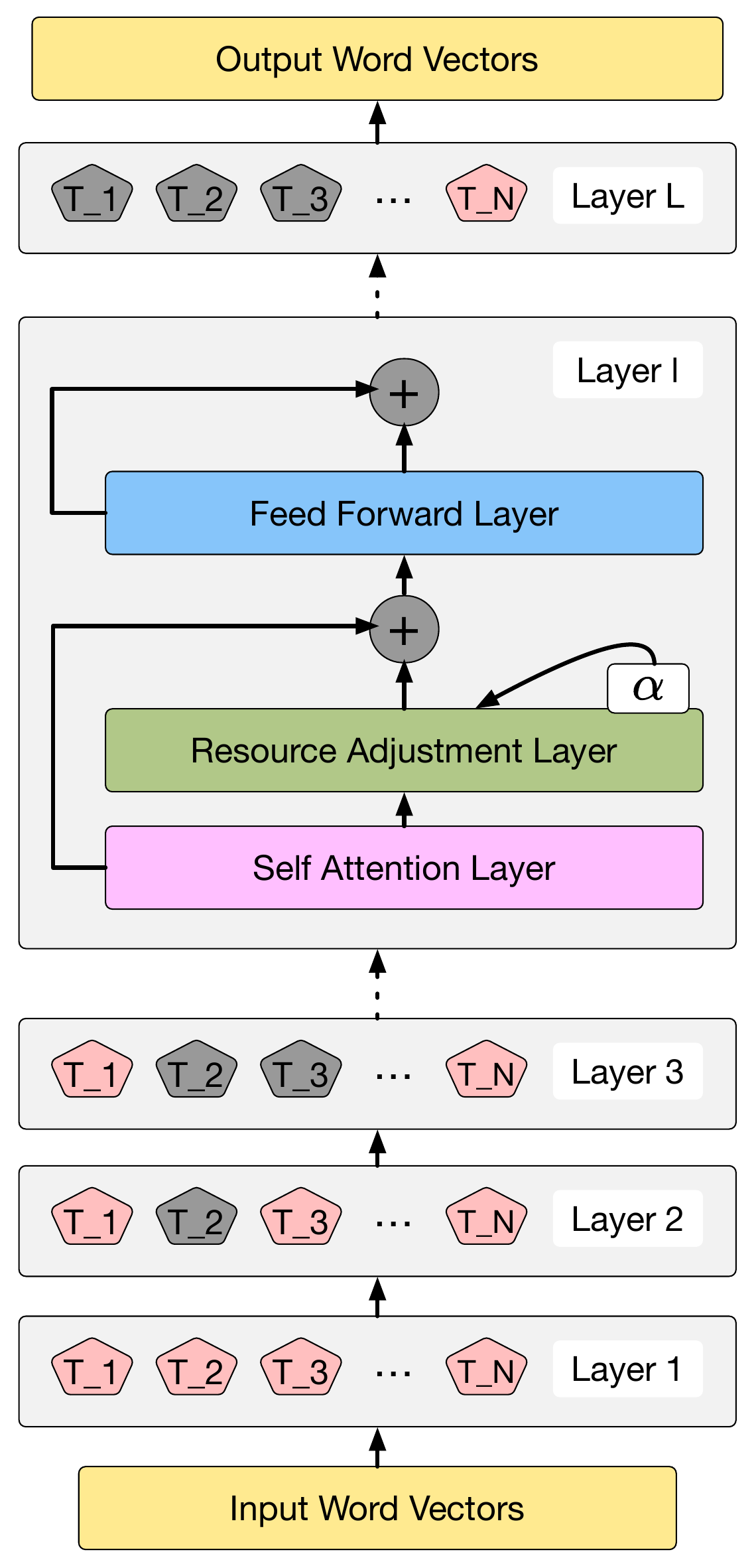}}
\caption{The proposed dynamic resource attention architecture.}
\label{fig1}
\end{figure}

As discussed in the previous section, the soft-score determines the contribution of each hidden activation to the layer output. We assume that hidden activations with lower contributions can be eliminated in subsequent layers when inference acceleration is required. However, prior approaches face key bottlenecks, including maintaining stable inference performance, enabling efficient zero-shot acceleration, and ensuring a broad and reliable acceleration range. Figure~\ref{fig1} illustrates the proposed dynamic resource attention architecture.

As shown in Figure~\ref{fig1}, the proposed architecture eliminates hidden activations with lower contributions to the attention layer output, based on the desired inference acceleration. This process is governed by the hyperparameter $\alpha$, which controls the proportion of preserved activations. The number of remaining hidden activations in layer $l$ is defined as:

\begin{equation}
N_l = \max(\alpha N_{l-1}, \tau_l)
\label{eq2}
\end{equation}

Here, $N_l$ represents the number of preserved hidden activations at layer $l$, $N_{l-1}$ is the number of activations from the previous layer, and $\alpha$ is the preservation rate hyperparameter, a floating-point value between 0 and 1. Reducing $\alpha$ decreases the number of tokens retained per layer. For example, setting $\alpha = 0.3$ means that at each Transformer layer, only 30\% of hidden activations are preserved while the remaining 70\% are pruned.

The parameter $\tau_l$ plays a critical role in maintaining model stability. It sets a minimum threshold on the number of preserved activations, preventing extreme cases where only a single token survives in the attention layer—a scenario that would destabilize the model. This clipping mechanism not only improves stability during fine-tuning but also enhances inference performance when operating under high pruning rates.

\subsection{Zero-Shot Adjustable Acceleration} \label{sec3.4}
Optimizing acceleration after fine-tuning and during inference is essential for designing efficient architectures, a process referred to as zero-shot acceleration. Fine-tuning in large language models (LLMs) is inherently progressive, with the model gradually adapting to a downstream task. Prior studies show that applying a fixed preservation rate during fine-tuning often results in performance degradation when that rate is later altered during zero-shot acceleration. Conversely, scheduling and varying the preservation rate during fine-tuning causes the model to overfit to the most recently observed rates due to its stepwise adaptation process.

To address this limitation, the proposed method samples the preservation rate from a uniform distribution at each fine-tuning step:

\begin{equation}
\alpha = U(0.02, 1.00)
\label{eq3}
\end{equation}

By employing this distribution, the model is exposed to dynamic preservation rates ranging from 2\% to 100\% token preservation per layer. This strategy enhances robustness to preservation rate variations during zero-shot acceleration. Importantly, the proposed fine-tuning scheme fully replaces conventional fine-tuning without incurring additional training cost.

\section{Experiments}
As outlined in Section~\ref{sec3.1}, recent large language models (LLMs) exhibit substantial architectural similarities. To comprehensively evaluate the proposed method, we select GPT-2 as a legacy model representing earlier generations of LLMs, and LLaMA 3 and Gemma 2 as recently introduced state-of-the-art models. The effectiveness of the proposed Zero-Shot Adjustable Acceleration method is assessed across text classification, text generation, and instruction-tuning tasks, using a diverse set of benchmarks and datasets.

\subsection{Datasets}
The proposed method is evaluated on three categories of tasks: natural language understanding, text generation, and instruction tuning.

For natural language understanding, we use the IMDB dataset \cite{maas2011learning} for sentiment analysis and the GLUE benchmark \cite{wang2018glue}, which includes diverse tasks such as single-sentence classification (SST-2), sentence similarity (MRPC, QQP), question answering (QNLI), and natural language inference (RTE, MNLI-M, MNLI-MM).

For text generation, we adopt WikiText-103 \cite{merity2016pointer}, Penn Treebank (PTB) \cite{marcus1993building}, and One Billion Word (1BW) \cite{chelba2014billion} as primary benchmarks for language modeling. We also include the LAMBADA dataset \cite{paperno-EtAl:2016:P16-1}, which requires long-range contextual reasoning to predict the final word in narrative passages.

For instruction tuning, we use the Massive Multitask Language Understanding (MMLU) benchmark \cite{hendrycks2020measuring}, which consists of exam-style questions covering 57 subjects, including mathematics, history, law, and medicine.

Together, these datasets provide a comprehensive evaluation framework, spanning classification, generation, and reasoning tasks, and ensuring that the proposed method is tested across diverse application scenarios.

\subsection{Model Settings}
For GPT-2, LLaMA 3, and Gemma 2, the attention layer in each decoder is modified according to the proposed architecture described in Section~\ref{sec3.3}. The hyperparameter $\alpha$ controls the hidden activation preservation rate, enabling adjustable acceleration.

Evaluation is conducted with the following hyperparameters: a learning rate of \num{4e-5}, a batch size of 8, 5 training epochs, and a dropout rate of 0.1. The preservation rate $\alpha$ is varied from 0.02 to 1.00 for GPT-2 and from 0.50 to 1.00 for LLaMA 3 and Gemma 2 (see Section~\ref{sec3.4}). The minimum number of preserved hidden activations ($\tau_l$) is fixed at 5 to ensure stability, as discussed in Section~\ref{sec3.3}.

These configurations establish a consistent experimental setup for evaluating the effectiveness and robustness of the proposed method across both legacy and state-of-the-art LLMs.

\subsection{Implementation Details}
This study employs pre-trained GPT-2, LLaMA 3, and Gemma 2 models from the Hugging Face\footnote{\url{https://huggingface.co/}} library, implemented in PyTorch\footnote{\url{https://pytorch.org/}} and executed on an NVIDIA Tesla T4 GPU. Datasets are obtained from the Hugging Face Datasets library, and GLUE results are validated using the official GLUE evaluation service\footnote{\url{https://gluebenchmark.com/}}, ensuring consistency and reliability in performance reporting.

To fine-tune LLaMA 3 and Gemma 2 efficiently, the fine-tuning policy described in Section~\ref{sec3.4} is combined with the BitsAndBytes library~\cite{dettmers2023qlora}, which enables resource-efficient deployment of large models via k-bit quantization. Further efficiency gains are achieved through Low-Rank Adaptation (LoRA)~\cite{hu2021lora}. Together, these techniques significantly reduce the computational and memory overhead associated with fine-tuning large-scale LLMs.

\subsection{Results}
Table~\ref{table2} reports the results of the proposed Zero-Shot Adjustable Acceleration method applied to GPT-2 on the IMDB dataset and the GLUE benchmark. Reducing the $\alpha$ value yields substantial improvements in inference speed, achieving up to an 11× speedup on IMDB and an average latency reduction of approximately 9× across GLUE tasks. Since sequence lengths vary across datasets, the preservation rate is adjusted individually for each dataset and acceleration configuration.

These findings confirm the effectiveness of the proposed fine-tuning policy and demonstrate that inference latency can be dynamically controlled without requiring additional fine-tuning. Due to restrictions on the GLUE test server, GLUE results are reported on the validation sets.

\begin{table*}
\centering
\begin{tabular}{ccccccccc}
\hline

\textbf{Speedup} &
\textbf{IMDB} &
\textbf{MNLI\textsubscript{m}} &
\textbf{MNLI\textsubscript{mm}} &
\textbf{RTE} &
\textbf{QQP} &
\textbf{QNLI} &
\textbf{SST-2} &
\textbf{MRPC} \\

\hline

1x & 92.40 & 80.55 & 81.26 & 63.54 & 84.16 & 85.06 & 89.91 & 83.53 \\
2x & 92.34 & 80.53 & 81.25 & 64.26 & 84.16 & 85.06 & 89.56 & 83.78 \\
3x & 92.16 & 80.44 & 80.77 & 63.18 & 84.16 & 84.86 & 88.88 & 83.19 \\
5x & 91.13 & 76.42 & 77.79 & 62.45 & 80.10 & 81.91 & \textbf{82.00} & \textbf{76.16} \\
7x & 90.53 & \textbf{74.56} & \textbf{74.97} & 63.18 & 80.17 & 78.8 & - & - \\
9x & 90.05 & - & - & \textbf{56.32} & \textbf{77.74} & \textbf{69.63} & - & - \\
11x & \textbf{86.48} & - & - & - & - & - & - & - \\

\hline
\end{tabular}
\caption{
Experimental results of the proposed Zero-Shot Adjustable Acceleration method applied to GPT-2 (137M parameters) on the IMDB dataset~\citep{maas2011learning} and the GLUE benchmark~\citep{wang2018glue}. Performance is reported as F1-score for MRPC and QQP, and accuracy for all other tasks. Bold values denote the optimal speed–accuracy trade-off, where the model preserves full functionality.
}
\label{table2}
\end{table*}

\begin{figure*}[t]
\centerline{\includegraphics[width=1.0\linewidth]{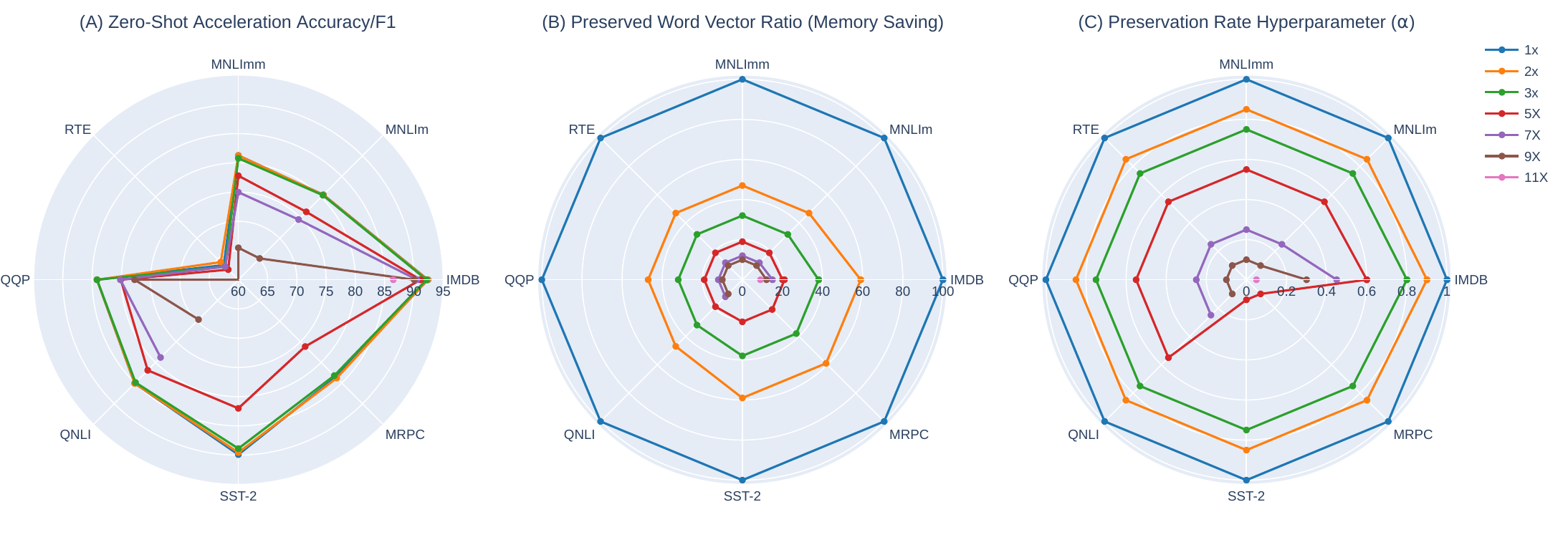}}
\caption{
Visualization of GPT-2 (137M parameters) performance across three dimensions: (A) zero-shot acceleration results, (B) ratio of preserved hidden activations, and (C) preservation rate hyperparameter ($\alpha$) over a 1×–11× acceleration range. Performance is reported as F1-score for MRPC and QQP~\citep{wang2018glue}, and accuracy for all other tasks.
}
\label{fig2}
\end{figure*}

Fig.\ref{fig2} illustrates the performance of the proposed method for GPT-2 across three dimensions: (A) Zero-Shot Acceleration results, demonstrating the method’s stability and effectiveness; (B) the ratio of preserved hidden activations, highlighting substantial memory savings; and (C) the range of the preservation rate hyperparameter ($\alpha$), showing enhanced effectiveness on datasets with longer sequence lengths. Additional visualizations of hidden activation preservation are provided in Appendix\ref{hidden_visualization}.

Table~\ref{table3} reports perplexity (PPL) results for GPT-2 on the WikiText-103, LAMBADA, PTB, and 1BW datasets. The proposed method demonstrates consistent effectiveness across different acceleration settings, achieving up to a 3× improvement in Time-to-First-Token (TTFT). These results indicate that the input sentence context is well preserved, supporting the validity of the hidden activation pruning strategy.

\begin{table}
\centering
\begin{tabular}{ccccc}
\hline

\textbf{\thead{TTFT \\ Speedup}} &
\textbf{\thead{Wiki \\ Text-103}} &
\textbf{\thead{LAMBADA}} &
\textbf{\thead{PTB}} &
\textbf{\thead{1BW}} \\

\hline

1x & 34.48 & 52.14 & 38.69 & 48.89 \\
2x & 39.10 & 55.56 & 46.40 & 49.20 \\
3x & 48.59 & 62.16 & 59.07 & 52.30 \\

\hline
\end{tabular}
\caption{
Perplexity (PPL) results for GPT-2 (137M parameters) on the WikiText-103~\citep{merity2016pointer}, LAMBADA~\citep{paperno-EtAl:2016:P16-1}, PTB~\citep{marcus1993building}, and 1BW~\citep{chelba2014billion} datasets. The 1BW result is evaluated on 0.5\% of the test set.
}
\label{table3}
\end{table}

Table~\ref{table4} reports the performance of the proposed method on LLaMA 3 and Gemma 2, evaluated using the MMLU benchmark. The method effectively reduces Multiply-Accumulate Operations (MACs) during inference while maintaining high accuracy, demonstrating its applicability to both legacy and state-of-the-art models. In addition, the GPU memory footprint reduction (excluding constant self-model memory) and the preservation rate ($\alpha$) are reported across the zero-shot acceleration range. Since both models are quantized and executed on GPUs, MACs provide a more reliable measure of performance than Time-to-First-Token (TTFT), given the characteristics of GPU acceleration.

\begin{table*}
\centering
\begin{tabular}{c|ccc|ccc}
\hline

Model &
\multicolumn{3}{c}{\textbf{LLaMA-3-8B-4bit}} &
\multicolumn{3}{c}{\textbf{Gemma-2-2B-4bit}} \\

\thead{MACs \\ Reduction} & \textbf{\thead{MMLU \\ Score}} & \thead{GPU \\ Footprint} & \thead{Preservation \\ Rate ($\alpha$)} 
& \textbf{\thead{MMLU \\ Score}} & \thead{GPU \\ Footprint} & \thead{Preservation \\ Rate ($\alpha$)} \\

\hline

\textbf{1x}   & 62.95 & 1.00 & 1.00 & 60.21 & 1.00 & 1.00 \\
\textbf{1.5x} & 58.28 & 1.70 & 0.99 & 54.60 & 1.67 & 0.99 \\
\textbf{2.5x} & 54.32 & 2.05 & 0.98 & 49.15 & 1.70 & 0.97 \\
\textbf{3.5x} & 49.95 & 2.72 & 0.97 & 45.51 & 1.71 & 0.96 \\

\hline
\end{tabular}
\caption{
Zero-shot acceleration results showing the reduction in Multiply-Accumulate Operations (MACs) achieved by the proposed approach on LLaMA 3~\cite{dubey2024llama} and Gemma 2~\cite{team2024gemma2}, evaluated on the MMLU~\cite{hendrycks2020measuring} instruction-tuning dataset. For each model, the MMLU score, GPU memory footprint reduction (excluding constant self-model memory), and preservation rate ($\alpha$) are reported across the zero-shot acceleration range.
}
\label{table4}
\end{table*}

Based on the results in Table~\ref{table4}, the proposed fine-tuning policy demonstrates strong performance when combined with the LoRA-based acceleration method. Furthermore, because quantized models are more sensitive to structural variations, the method is also effective with 4-bit quantized models, providing both enhanced acceleration and maintained performance. These findings confirm that the proposed approach does not alter the core model architecture and can be seamlessly integrated with other acceleration techniques.

It is important to note that, as with all pruning techniques, acceleration is typically accompanied by some degradation in accuracy or model quality, and the proposed method is no exception. In our approach, the model disruption ratio depends on the sample length and the type of context or task. For instance, samples with low redundancy and complex concepts experience greater accuracy degradation under acceleration compared to simpler, less informative samples.

\section{Conclusion}
This paper reviews recent advancements in large language model (LLM) architectures and introduces a novel acceleration framework. The proposed fine-tuning policy ensures stable performance across a wide range of zero-shot acceleration settings. Termed Zero-Shot Adjustable Acceleration, this approach enables dynamic adjustment of inference speed without requiring additional fine-tuning. Experimental results on GPT-2, LLaMA 3, and Gemma 2 demonstrate that the method is effective for both legacy and state-of-the-art LLMs, achieving up to 11× acceleration while maintaining high performance. These findings highlight that, although the full network is essential during initial training, selective pruning of hidden activations can substantially improve efficiency with minimal degradation in model quality.

\section{Limitations}

While the proposed Zero-Shot Adjustable Acceleration method demonstrates substantial improvements in inference speed and hardware utilization, several limitations should be considered:

\begin{enumerate}
\item \textbf{Task-Specific Performance Variability:} The effectiveness of the approach may vary across tasks. Although up to an 11× speedup is observed in some scenarios, tasks requiring complex reasoning or high accuracy may experience more pronounced performance degradation under increased acceleration.

\item \textbf{Model Compatibility:} The method is primarily designed for newly developed models. Its applicability to existing large language models or those with specialized architectures requires further investigation.  
\item \textbf{Code Availability:} To ensure reproducibility and facilitate further research, we plan to release the code and trained models upon publication. This will enable the community to validate our findings and adapt the approach to other models and tasks.  
\item \textbf{Evaluation Scope:} Experiments focus on classification and text generation tasks. The generalizability of the method to other domains, such as multimodal applications or real-time systems, remains to be explored.  
\item \textbf{Resource Requirements:} Although the approach reduces computational demands during inference, the initial fine-tuning phase still requires substantial resources, which may limit accessibility for researchers with constrained hardware.  

\end{enumerate}

\bibliography{custom}
\bibliographystyle{acl_natbib}

\appendix

\section{Comparative Review of LLM Architectures}
\label{llm_comparisons}

A review of recent Large Language Model (LLM) architectures—including LLaMA 2~\citep{touvron2023LLaMA2}, LLaMA 3~\citep{dubey2024llama}, Gemma 1~\citep{team2024gemma}, Gemma 2~\citep{team2024gemma2}, Mistral~\citep{jiang2023mistral}, and the earlier GPT-2~\citep{radford2019language}—reveals that, despite differences in training scale, dataset composition, and optimization strategies, these models share notable architectural commonalities. Key areas of similarity include:

\begin{itemize}
\item \textbf{Overall Transformer Backbone:} All models follow either the encoder-decoder or decoder-only Transformer paradigm, with stacked self-attention and feed-forward layers.
\item \textbf{Activation Functions:} Variants of GELU, SwiGLU, or similar smooth nonlinearities are consistently employed.
\item \textbf{Residual Connections and Normalization:} Residual pathways coupled with LayerNorm—often in pre-norm configurations—are universal across these architectures.
\item \textbf{Attention Mechanisms:} Multi-head self-attention, with minor implementation variations (e.g., rotary position embeddings or grouped-query attention), is a shared foundation.
\end{itemize}

As summarized in Table~\ref{table1}, models such as GPT-2, Gemma 2, Mistral, and LLaMA 3 exhibit particularly strong alignment in structural design. These observations suggest that modern LLMs are converging toward a relatively standardized architectural blueprint. This convergence indicates that developing a universal acceleration and optimization platform capable of supporting a wide range of Transformer-based models is both technically feasible and strategically advantageous. Such a platform would facilitate broader applicability of hardware and software optimizations, streamline deployment pipelines, and enhance research reproducibility across diverse model families.

\begin{table*}
\centering
\begin{tabular}{cccccccc}
\hline
\textbf{\thead{Model \\ Name}} & 
\textbf{\thead{Architecture \\ Type}} & 
\textbf{\thead{Model \\ Size}} &
\textbf{\thead{Number of \\ Layers/Heads}} &
\textbf{\thead{Hidden \\ Size}} &
\textbf{\thead{Activation \\ Function}} &
\textbf{\thead{Residual \\ Branch Type}} &
\textbf{\thead{Attention \\ Type}} \\
\hline

GPT-2 & Decoder & 355M & 24/16 & 1024 & GeLU & Pre-Norm & MHA \\
Gemma 2 & Decoder & 2B & 26/8 & 2304 & SwiGLU & Pre-Norm & MQA \\
Mistral & Decoder & 7B & 32/32 & 4096 & SwiGLU & Pre-Norm & GQA \\
LlaMA 3 & Decoder & 8B & 32/32 & 4096 & SwiGLU & Pre-Norm & GQA \\

\hline
\end{tabular}
\caption{
Comparative summary of the architectures of GPT-2~\citep{radford2019language}, Gemma 2~\citep{team2024gemma2}, Mistral~\citep{jiang2023mistral}, and LLaMA 3~\citep{dubey2024llama}.
}
\label{table1}
\end{table*}

\section{Hidden Activation Preservation Visualization}
\label{hidden_visualization}

\begin{figure*}[t]
\centerline{\includegraphics[width=1.0\linewidth]{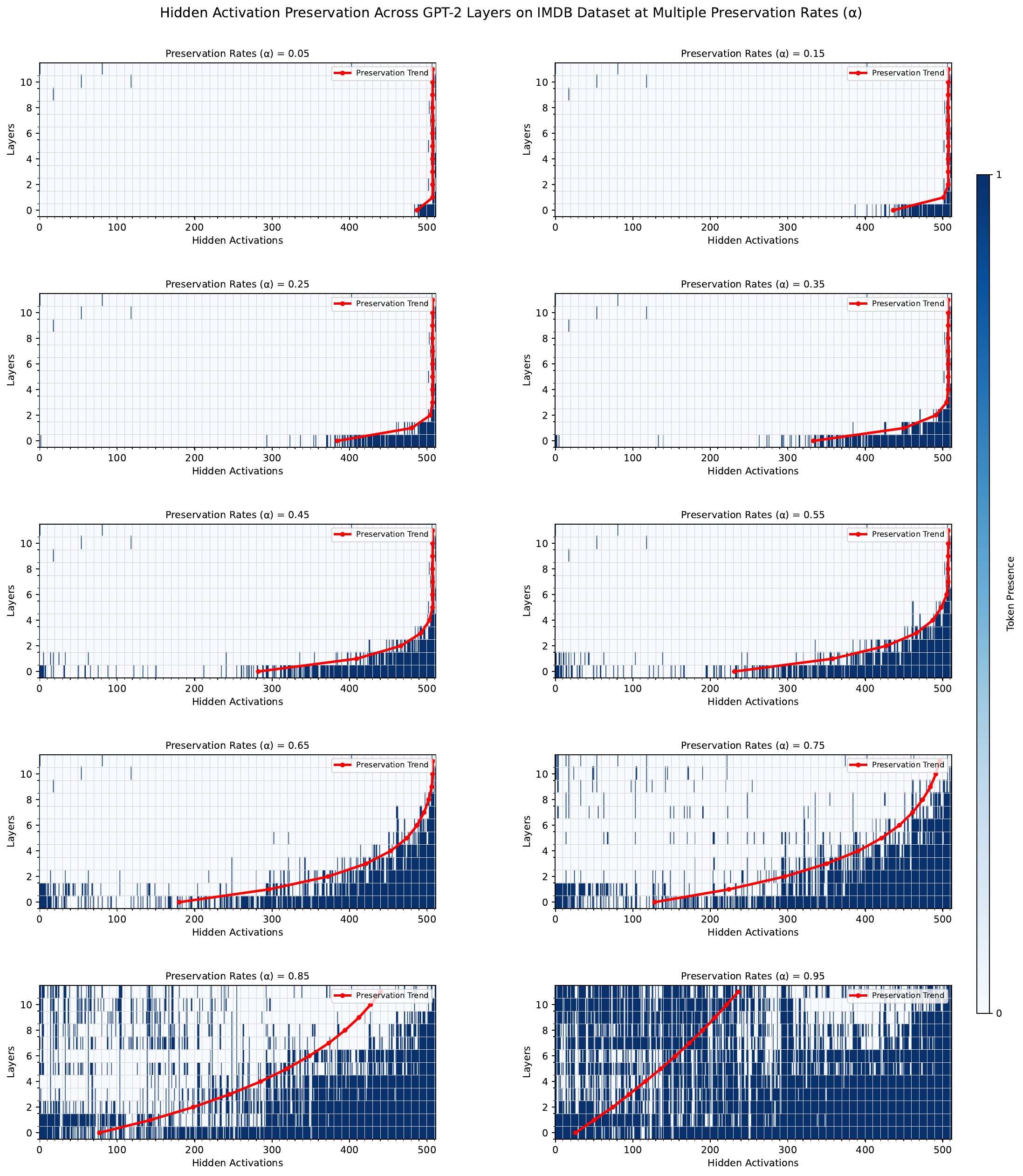}}
\caption{
Binary visualization of hidden activation preservation across GPT-2 layers on the IMDB~\citep{maas2011learning} dataset at different preservation rates ($\alpha$). Blue squares represent preserved tokens, white squares represent pruned tokens, and the red line indicates the overall preservation trend per layer.
}
\label{fig3}
\end{figure*}

Fig.~\ref{fig3} visualizes hidden activations in GPT-2 across multiple preservation rates ($\alpha$) during inference on the IMDB dataset. Each subplot corresponds to a different preservation rate, showing preserved activations (blue) versus pruned activations (white) at each layer.

For clarity, all hidden activations are displayed, including those marked as pruned but not removed. This allows some activations to be reactivated in later layers, which is an exceptional case compared to the method description.

The red trend line in each subplot indicates the total number of preserved hidden activations per layer. Since GPT-2 uses causal self-attention, each hidden activation depends on preceding activations in the sequence. As the preservation rate decreases, the trend line becomes steeper, reflecting increased pruning per layer, with pruning effects generally accumulating toward the end of the sequence. This behavior corresponds to a reduction in computational load, accelerating inference while preserving essential information.

Overall, the figure demonstrates how selective preservation of hidden activations varies across layers and preservation rates, providing insight into the layer-wise dynamics of token pruning and its effect on model efficiency.

\end{document}